\documentclass[sigconf]{acmart}
\usepackage{multirow}
\AtBeginDocument{%
  }

\setcopyright{acmlicensed}
\copyrightyear{2025}
\acmYear{2025}
\acmDOI{XXXXXXX.XXXXXXX}
\acmISBN{978-1-4503-XXXX-X/25/11}

\begin{document}

%%
%% The "title" command has an optional parameter,
%% allowing the author to define a "short title" to be used in page headers.
\title{ Self-Evolving LLMs via Continual Instruction Tuning}
% \title{Continual Instruction Tuning of LLMs via \\ Mixture of LoRA Experts}

%%
%% The "author" command and its associated commands are used to define
%% the authors and their affiliations.
%% Of note is the shared affiliation of the first two authors, and the
%% "authornote" and "authornotemark" commands
%% used to denote shared contribution to the research.

\author{Jiazheng Kang}
\affiliation{%
  \institution{Beijing University of Posts and Telecommunications}
  \streetaddress{}
  \city{Beijing}
  \country{China}}
\email{kjz@bupt.edu.cn}

\author{Le Huang}
\affiliation{%
  \institution{Beijing University of Posts and Telecommunications}
  \streetaddress{}
  \city{Beijing}
  \country{China}}
\email{lehuang@bupt.edu.cn}

\author{Cheng Hou}
%\authornote{Corresponding author.}
\affiliation{%
  \institution{Tencent AI Lab}
  \streetaddress{}
  \city{Beijing}
  \country{China}
}
\email{chenghou@tencent.com}

\author{Zhe Zhao}
%\authornote{Corresponding author.}
\affiliation{%
  \institution{Tencent AI Lab}
  \streetaddress{}
  \city{Beijing}
  \country{China}
}
\email{nlpzhezhao@tencent.com}

\author{ZhenXiang Yan}
%\authornote{Corresponding author.}
\affiliation{%
  \institution{Tencent AI Lab}
  \streetaddress{}
  \city{Beijing}
  \country{China}
}
\email{kimmoyan@tencent.com}

% \author{Chuan Shi}
% % \authornote{Corresponding author.}
% \affiliation{%
%   \institution{Beijing University of Posts and Telecommunications}
%   \streetaddress{}
%   \city{Beijing}
%   \country{China}
% }
% \email{shichuan@bupt.edu.cn}

\author{Ting Bai*}
\affiliation{%
  \institution{Beijing University of Posts and Telecommunications}
  \streetaddress{}
  \city{Beijing}
  \country{China}}
\thanks{*Corresponding author.}
\email{baiting@bupt.edu.cn}

%%
%% By default, the full list of authors will be used in the page
%% headers. Often, this list is too long, and will overlap
%% other information printed in the page headers. This command allows
%% the author to define a more concise list
%% of authors' names for this purpose.
\renewcommand{\shortauthors}{Kang, et al.}

%%
%% The abstract is a short summary of the work to be presented in the
%% article.
\begin{abstract}

In real-world industrial scenarios, large language models (LLMs) require Continuous Learning (CL) to adapt to diverse tasks as operational requirements diversify, demanding self-evolution capabilities to autonomously refine their knowledge and adapt to dynamic environments. However, existing CL approaches, such as replay-based and parameter isolation techniques, struggle with the catastrophic forgetting problem: new task training degrades performance on prior tasks due to the model’s adaptation to new data distributions, which weakens its generalization to old tasks. To address this issue, we propose a novel parameter-efficient adversarial MoE framework, MoE-CL, for industrial-scale self-evolving continual instruction tuning of LLMs. Specifically, MoE-CL employs a dual-expert architecture to enable self-evolution: a dedicated LoRA expert for each task to preserve task-specific knowledge, ensuring parameter independence and mitigating forgetting, and a shared LoRA expert to facilitate cross-task knowledge transfer. Specifically, a task-aware discriminator within a Generative Adversarial Network (GAN) is integrated into the shared expert to suppress task-irrelevant noise, ensuring only task-aligned knowledge is transferred during sequential task training. Through adversarial training, the shared expert learns generalized representations that mimic the task-aware discriminator, while dedicated experts retain task-specific details, balancing knowledge retention and cross-task generalization—key to the model’s self-evolution by autonomously optimizing knowledge integration across tasks. Extensive experiments on a public MTL5 benchmark and an industrial Tencent3 benchmark validate MoE-CL’s effectiveness in self-evolving continual learning. In real-world A/B testing on content compliance review in the Tencent Video Platform, MoE-CL reduced manual review costs by 15.3\%, demonstrating its applicability for large-scale industrial deployment where self-evolution is critical for adapting to evolving operational demands. Implementation code is publicly available at \href{https://github.com/BAI-LAB/MoE-CL}{https://github.com/BAI-LAB/MoE-CL}.

\end{abstract}

%%
%% The code below is generated by the tool at http://dl.acm.org/ccs.cfm.
%% Please copy and paste the code instead of the example below.
%%


%%
%% Keywords. The author(s) should pick words that accurately describe
%% the work being presented. Separate the keywords with commas.
%% A "teaser" image appears between the author and affiliation
%% information and the body of the document, and typically spans the
%% page.
% \begin{teaserfigure}
%   \includegraphics[width=\textwidth]{sampleteaser}
%   \caption{Seattle Mariners at Spring Training, 2010.}
%   \Description{Enjoying the baseball game from the third-base
%   seats. Ichiro Suzuki preparing to bat.}
%   \label{fig:teaser}
% \end{teaserfigure}

% \received{20 February 2007}
% \received[revised]{12 March 2009}
% \received[accepted]{5 June 2009}

%%
%% This command processes the author and affiliation and title
%% information and builds the first part of the formatted document.
\maketitle

\section{Introduction}

In the era of large-scale industrial AI deployment, self-evolution—the ability of large language models (LLMs) to autonomously refine knowledge integration, adapt to dynamic task patterns, and retain prior competencies without external intervention~\cite{gao2025survey} has become indispensable. Industrial scenarios demand LLMs to rapidly respond to evolving operational demands across diverse real-world tasks: for instance, Tencent’s content compliance ecosystem, central to ensuring regulatory adherence and user safety, handles over two hundred thousand daily text reviews from fields like social platforms, news media, and e-commerce. Each domain presents distinct linguistic patterns, making self-evolution in continuous learning critical—only through autonomous adaptation can LLMs sustain performance across shifting tasks without constant human oversight. This operational complexity underscores the need for self-evolving continuous learning (CL) frameworks~\cite{wu2024continual, zheng2025towards, wang2024comprehensive}, which empower LLMs to seamlessly integrate new knowledge via iterative tuning and adapt dynamically to evolving data distributions.
However, self-evolution in continuous learning of LLMs is inherently challenged by the catastrophic forgetting problem. This phenomenon arises from the optimization dynamics of deep learning, where parameter updates during new task training unintentionally disrupt or overwrite the neural representations acquired from prior tasks. As a result, the model suffers significant performance degradation on previously mastered tasks, as the iterative fine-tuning for new tasks undermines the stable knowledge encoding necessary for maintaining competence in previous tasks—a critical barrier to self-evolution, which requires autonomous retention of old knowledge while integrating new insights.

Existing solutions in continuous learning of LLMs fail to fully enable self-evolution, facing critical trade-offs between knowledge retention and new task adaptation. 
For example, replay-based methods~\cite{rebuffi2017icarl,guo2024corpusbrain++, sun2019lamol} preserve prior knowledge by reusing historical data or generating pseudo-samples, but these approaches suffer from data contamination: synthetic data often introduces noise that distorts task-specific representations. Besides, the computational overhead of storing and processing replay data renders them impractical for large-scale industrial use. 
Regularization techniques~\cite{aljundi2018memory, kirkpatrick2017overcoming, huang2021continual} constrain parameter updates to protect important weights, yet they overly restrict the model’s ability to specialize in new tasks with distinct requirements, limiting adaptability to operational demands in industry scenarios. 
On the other hand, dynamic architecture approaches, like parameter isolation techniques~\cite{rusu2016progressive, ke2021adapting, luo2023mitigating},  allocate dedicated parameters to each task, effectively preventing inter-task interference and retaining old task performance. However, such an isolationist design limits cross-task knowledge transfer, failing to leverage shared semantic patterns (e.g., common semantic features across tasks with related content information).
In real-world AI deployments, this oversight limits the model’s ability to generalize across domains and accumulate fine-tuning gains, leading to suboptimal scalability when handling diverse sequential tasks, which thus constitutes a major impediment to self-evolution in LLMs that rely on autonomous knowledge integration across tasks.

% Existing strategies, whether prioritizing knowledge retention at the cost of computational efficiency or relying on parameter isolation that hinders cross-task generalization, fail to achieve the balance required for LLMs to be applied in complex industrial environments with evolving operational demands.

In our work, we introduce \textbf{MoE-CL}, a novel adversarial \textbf{Mixture of LoRA Experts} (\textbf{MoE}) architecture designed for self-evolving \textbf{Continuous Learning} (\textbf{CL}) in large language models. Our core objective is to enable LLMs to autonomously transfer knowledge from previously learned tasks to new ones during sequential training while minimizing the impact of new task updates on old task performance, a challenge that captures the essence of self-evolution of continuous learning in large-scale industrial deployments of LLMs. 
MoE-CL addresses catastrophic forgetting and enables useful knowledge transfer through an adversarial LoRA expert architecture. By allocating a dedicated LoRA expert to each task, MoE-CL ensures that training a new task does not overwrite prior parameters, inherently supporting autonomous knowledge retention. Concurrently, the shared LoRA expert serves as a self-optimizing cross-task bridge, learning generalized representations that capture common semantic patterns across tasks. Specifically, the collaborative mechanism between experts is enhanced via a Generative Adversarial Network (GAN): a task-aware discriminator suppresses task-irrelevant noise in the shared expert, ensuring only task-aligned knowledge is transferred, which enables the model to autonomously refine knowledge integration without external guidance.
During inference, MoE-CL adaptively combines outputs from the shared LoRA expert and the specific LoRA expert for the fine-tuning of the new task. 
MoE-CL freezes the parameters of other tasks and only updates the parameters in the task-specific LoRA and the shared LoRA expert. By doing so, it minimizes the computational overhead, which makes it well-suited for self-evolving large-scale industrial systems. The contributions of our paper are summarized as follows:

\begin{itemize}
\item We propose a novel adversarial mixture of LoRA experts architecture (MoE-CL) for self-evolving continual instruction tuning of LLMs. MoE-CL achieves self-evolution by maintaining parameter independence through dedicated LoRA experts and integrating common knowledge via a shared LoRA expert, thus addressing the catastrophic forgetting problem.

\item We design a task-aware discriminator in a generative adversarial network, which enhances the self-evolving capability of MoE-CL. It enables the model to autonomously transfer task-relevant knowledge while suppressing task-irrelevant noise, thereby improving knowledge generalization in continual instruction tuning.

\item Extensive experiments on the public MTL5 benchmark and industrial Tencent3 benchmark demonstrate that MoE-CL outperforms state-of-the-art baselines, with its self-evolving ability validated by consistent performance across diverse tasks. In particular, an offline A/B test on content compliance review in the Tencent Video Platform shows a 15.3\% improvement in stripping rate, confirming its practical feasibility for large-scale industrial deployments requiring dynamic self-adaptation.
\end{itemize}

\section{Related Work}
Our MoE-CL architecture, a novel adversarial Mixture of LoRA Experts framework, connects with four related work areas, i.e., Self-Evolution of LLMs, Continual Learning, Continual Instruction Tuning, and Adversarial Learning with MoE. Our work addresses catastrophic forgetting and enables effective knowledge transfer in LLMs' continual instruction tuning for self-evolution.

\subsection{Self-Evolution of LLMs}
Self-evolution of LLMs is defined as the ability of models to autonomously adapt to dynamic tasks, integrate cross-task knowledge, and sustain performance without heavy external intervention~\cite{gao2025survey}. 
Existing approaches for enabling such capability generally fall into three categories: Autonomous Learning Mechanisms, which enable self-improvement via self-generated supervision~\cite{wu2024meta}, internal feedback~\cite{madaan2023self}, or rewarding signals~\cite{tao2024survey}; Dynamic Architecture Adaptation, which focuses on structural optimization, including modular design search~\cite{du2025survey}, workflow generation~\cite{ho2025polymath}, and fine-tuing architecture design~\cite{kong2025phasenas}; and Knowledge Integration Frameworks, which consolidate cross-task knowledge through memory management~\cite{wu2025human}, tool evolution~\cite{luo2025gate}, or domain-specific toolset creation~\cite{xu2025petoolllm}. 
Current research thus lacks a solution that balances autonomous knowledge retention and adaptive transfer, which are core challenges to self-evolution in continuous instruction tuning. 
Our work 
%falls within the category of dynamic architecture adaptation: 
MoE-CL, a Mixture of LoRA Experts architecture, features dedicated experts for autonomous knowledge preservation, a shared expert for cross-task integration, and a GAN-based discriminator for noise suppression, achieving LLM self-evolution through adaptive continual instruction tuning to adapt to sequential task dynamics.

% Our work falls in dynamic architecture adaptation,  Our work addresses this gap with MoE-CL, a Mixture of LoRA Experts architecture featuring dedicated experts for autonomous knowledge preservation, a shared expert for cross-task integration, and a GAN-based discriminator for noise suppression, achieving self-evolution of LLMs through adaptive continual instruction tuning, as a representative of Dynamic Architecture Adaptation.

\subsection{Continual Learning}

Continual learning (CL), or termed as lifelong learning in large language models, plays a critical role in overcoming the limitations of traditional static-dataset training. The aim of continual learning is to incrementally incorporate new knowledge, adapt to diverse tasks across evolving domains, and retain previously acquired capabilities throughout the learning process. 
By enabling incremental learning across shifting domains and diverse tasks, continual learning ensures that LLMs not only adapt to emerging information but also maintain foundational competencies, directly addressing the critical challenge of "catastrophic forgetting" inherent in static training and allowing them to remain relevant and effective in dynamic real-world scenarios where knowledge and tasks evolve continuously. The approaches in CL can be generally categorized into three types~\cite{zheng2025towards} based on their knowledge integration mechanisms: continual pre-training, continual fine-tuning, and external knowledge integration (including retrieval-based and tool-based methods).
Continual pre-training enhances LLMs' knowledge more efficiently than full pre-training by incrementally incorporating new data streams, enabling cross-domain generalization without significant computational overhead~\cite{gupta2023continual, ke2023continual}.
For example, ELLE~\cite{ELLE} uses function-preserved model expansion and pre-trained domain prompts to continuously incorporate streaming data and enhance performance.
Continual fine-tuning adapts LLMs to specific tasks while preserving prior expertise through techniques, such as contrastive ensemble distillation for text classification~\cite{classic},  incremental knowledge transfer for named entity recognition~\cite{ExtendNER}, and constrained optimization to balance new task priorities~\cite{cppo}.
% Continual fine-tuning adapts LLMs to specific tasks. For example, CLASSIC \cite{classic} uses contrastive ensemble distillation for text classification, ExtendNER \cite{ExtendNER} applies knowledge distillation for named entity recognition, and CPPO \cite{cppo} balances new priorities in value alignment.
External knowledge integration in CL bridges gaps in LLMs' internal representations via retrieval-based methods like optimized document retrieval~\cite{dpr} or tool-based approaches such as Chameleon~\cite{chameleon}, which integrates web search and Python functions. These techniques enhance real-time reasoning capabilities and extend model utility beyond pre-trained knowledge of LLMs.
% External knowledge integration in CL includes retrieval-based (e.g., DPR \cite{dpr} for optimized retrieval) and tool-based (e.g., Chameleon \cite{chameleon} for tool interaction) approaches to boost LLMs' capabilities.

\subsection{Continual Instruction Tuning}

Continual instruction tuning~\cite{razdaibiedina2023progressive, he2023continual, wang2024inscl} in LLMs refers to dynamically adapting LLMs through sequential task-specific instruction tuning, enabling them to incrementally incorporate new knowledge from evolving instructions and adapt to diverse tasks while retaining performance on previously learned tasks.
This paradigm aims to address the limitations of static training by facilitating iterative updates aligned with new instructional inputs, thereby enhancing the model’s flexibility in dynamic real-world scenarios.
A critical challenge in continual instruction tuning is the catastrophic forgetting problem, where parameter updates during new task training inadvertently disrupt or overwrite the neural representations acquired from prior tasks. 
This leads to significant performance degradation on previously learned tasks, as the model fails to preserve the stable knowledge encoding necessary for previous tasks. 

Existing approaches, such as replay-based, regularization, or architecture-based techniques, struggle to balance knowledge retention and new task adaptation.
For example, the replay-based method LAMOL~\cite{lamol} that uses generative replay to preserve past knowledge through pseudo-samples. It suffers from data noise and computational overhead. Regularization-based techniques such as ARPER \cite{arper} stabilize parameter updates via adaptive regularization. 
This type of method overly restricts model specialization, 
Architecture-based methods like TPEM \cite{geng2021continual} dynamically adjust network structures to retain task relevance. The isolated parameters limit cross-task knowledge transfer. 
The SOTA continual instruction tuning method is MoCL~\cite{wang-etal-2024-rehearsal}, which equips each task with a dedicated PEFT module and calculates the similarity between the input and the task vector as weights to fuse the outputs of the training task. However, it faces inherent limitations in balancing task specificity with cross-domain generalization.

Different from the above instruction tuning methods, we propose MoE-CL, which integrates dedicated task-specific experts (to preserve task-specific knowledge) and shared experts (to enable controlled knowledge transfer), mitigating catastrophic forgetting through adversarial training, alleviating task-irrelevant noise and improving efficient cross-task generalization.

\subsection{Adversarial Learning with MoE}
Learning with Generative Adversarial Networks (GAN) is a powerful technique that trains two models simultaneously: a generator that produces realistic outputs and a discriminator that distinguishes between real and generated data~\cite{goodfellow2020generative, creswell2018generative, goodfellow2014generative}. This adversarial process enhances the generator's ability to create outputs that are indistinguishable from real data, making it highly effective for tasks that require robustness and invariance.
In recommendation systems, adversarial learning has demonstrated effectiveness in addressing biases~\cite{wang2022invariant} and enhancing generalization by separating shared and task-specific features~\cite{bai2024efficient}. This ability to explicitly distinguish between generalizable knowledge and task-specific details aligns with the core challenges of continual learning in large language models, where there is a critical need to balance knowledge retention and adaptation to new tasks.
Building on the similar foundation of disentangling generalizable and task-specific knowledge, the Mixture-of-Experts (MoE) architecture has shown promise in analogous contexts. MoE enables models to dynamically combine specialized and shared knowledge, a strategy that has proven successful in traditional multi-task learning~\cite{ma2018modeling, tang2020progressive} and general LLM fine-tuning~\cite{li2024mixlora,dou2023loramoe}.  
However, despite these advancements, the potential of MoE in addressing the catastrophic forgetting problem in LLMs’ continual learning has remained underexplored.

Inspired by how adversarial learning can guide experts to learn more discriminative and task-aligned representations, our work bridges this gap by pioneering the integration of an adversarial MoE framework into LLMs’ continual instruction tuning. 
By leveraging adversarial training to regulate interactions between dedicated task-specific LoRA experts and a shared cross-task expert, we enable the model to retain task-specific knowledge while facilitating controlled knowledge transfer, thus addressing catastrophic forgetting in a novel way. Our approach makes a systematic exploration of MoE’s capabilities in this LLM continuous learning domain, leveraging the strengths of both adversarial learning and MoE architectures to enhance the continuous learning capacity of LLMs.

% In recommendation systems, adversarial learning helps address biases and improve generalization. For example, InvPref~\cite{wang2022invariant} uses adversarial learning to separate invariant preferences (true user preferences) from variant preferences (preferences influenced by latent biases) by confusing an environment classifier via a gradient reversal layer. In MPT-Rec~\cite{bai2024efficient}, adversarial learning separates general features shared across tasks from specific features unique to each task. This separation accelerates generalization for new tasks, making the recommendation system more adaptable and efficient.
% While the MoE architecture has been widely adopted in traditional multi-task learning~\cite{ma2018modeling, tang2020progressive, bai2022contrastive} and general LLM fine-tuning~\cite{li2024mixlora,dou2023loramoe}, our work pioneers its application in continual learning for LLMs to address catastrophic forgetting, marking the first exploration of MoE's potential in this critical domain.

% Inspired by the different types of experts learning through generative adversarial learning, our work pioneers its application in continual learning for LLMs to address catastrophic forgetting, marking the first exploration of MoE's potential in this critical domain.

\begin{figure*}
  \centering
  \includegraphics[width=\textwidth]{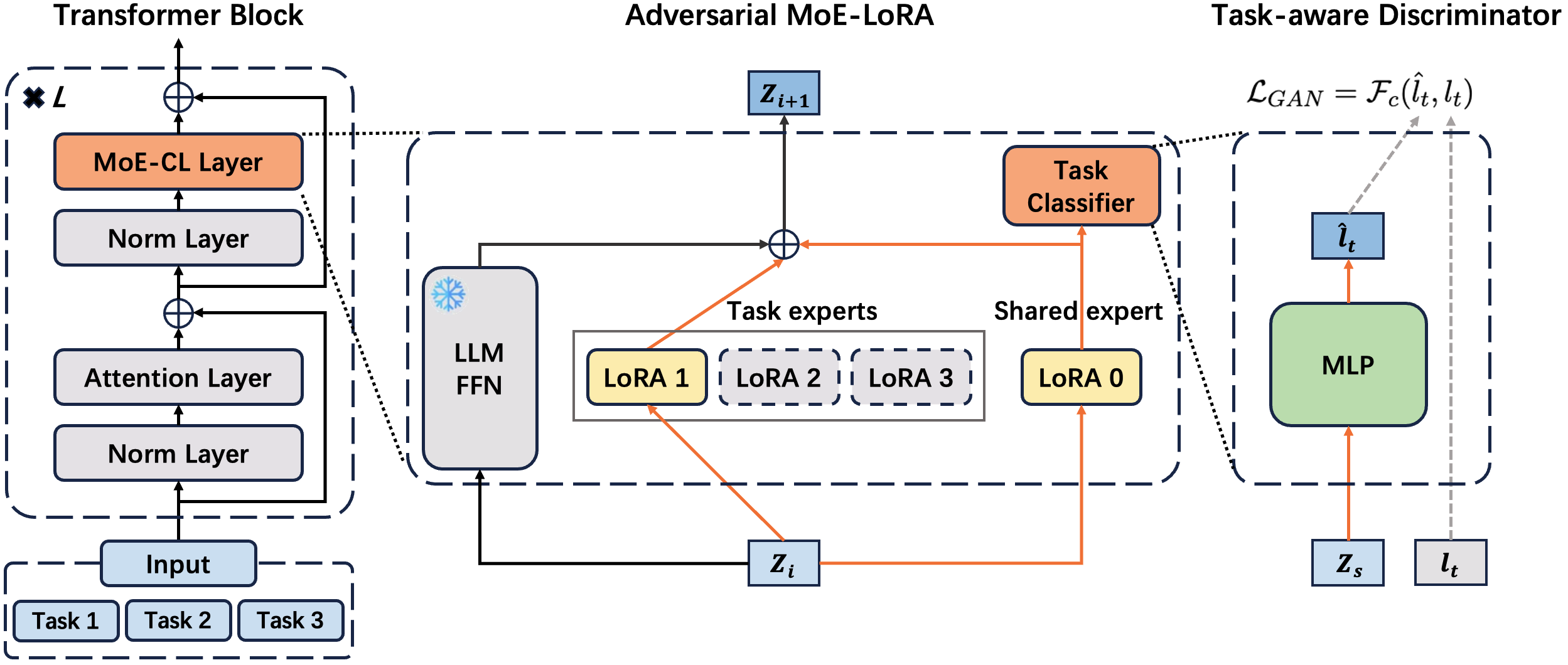}
  \caption{The overall architecture of MoE-CL: An adversarial MoE-LoRA framework integrating dedicated LoRA experts (for task-specific knowledge preservation), a shared LoRA expert (for cross-task knowledge transfer), and a GAN-based task-aware discriminator (to suppress task-irrelevant noise), which collectively alleviates catastrophic forgetting in continual instruction tuning of LLMs.
  }
  \label{fig:moecl_framework}
\end{figure*}

\section{Perliminary}
This section introduces the key concepts in our paper, including continual learning, instruction tuning with LoRA experts in the mixture-of-experts architecture.

\subsection{Continual Learning}

Continual learning (CL) enables models to sequentially acquire new tasks while retaining prior knowledge. Our work focuses on continual instruction tuning of LLMs with a series of tasks. 
Given a task sequence, represented as $\{T_1, ..., T_N\}$. Each task $T_i$ contains a set of learning samples $\{(x_i, y_i)\}$ for instruction tuning, where $x_i$ represents an input sample, $y_i$ is the corresponding ground truth label, and $ i \in \{1, ..., N\}$ represents the task identifier. 
The continual instruction tuning of LLMs aims to optimize the average performance of all the tasks after training all tasks sequentially. 

Let $\theta$ denote the parameters of the LLM. For each task $T_i$, we define a loss function $\mathcal{L}_i(\theta)$ that measures how well the model performs on the samples of that task. The overall goal of continual instruction tuning is to find the optimal parameter set $\theta$ that minimizes the average loss across all tasks in the sequence. It is formalized as:
\begin{equation}
\min_{\theta}\frac{1}{N}\sum_{i = 1}^{N}\mathcal{L}_i(\theta).
\end{equation}
By minimizing this average loss, we ensure that the LLM performs well not only on the most recently learned tasks but also on the entire set of tasks it has encountered during the sequential training process.

% Its overall objective is to optimize the average performance of the model across all tasks after learning all tasks sequentially.
% There is an additional objective that the performance loss of old tasks should be minimized after the training of new tasks. 

\subsection{Instruction Tuning with LoRA}
We adopt the Parameter-Efficient Fine-Tuning (PEFT) technique, i.e., LoRA~\cite{hu2022lora}, to allocate parameter updating for a specific task in the continual instruction tuning (i.e., supervised fine-tuning, SFT) of LLMs. 
LoRA adopts a low-rank decomposition technique to update the parameters of the decomposed parameter matrix to learn the data distribution of downstream tasks.
It modifies the Feed-Forward Neural Network (FFN) layers in Transformer blocks by introducing low-rank parameter updates. It freezes pre-trained parameters of LLMs while training adapter modules in FFN layers, enabling efficient fine-tuning of LLMs with minimal computational overhead. 
% For a linear layer in FFN represented as $\mathbf{h}=\mathbf{W}\mathbf{x}$, the update of the decomposed parameter matrix of LoRA is defined as:

For a linear layer in FFN, which is defined as $\mathbf{h}=\mathbf{W}\mathbf{x}$, the update of the decomposed parameter matrix of the LoRA is:
\begin{equation}
\mathbf{h} = \mathbf{W}\mathbf{x} + \Delta \mathbf{W} \mathbf{x} = \mathbf{W}\mathbf{x} + \frac{\alpha}{r} \mathbf{B}\mathbf{A}\mathbf{x},
\label{equ:lora}
\end{equation}
where the vector $\mathbf{x} \in \mathbb{R}^{I}$ encodes the input information, while $\mathbf{W} \in \mathbb{R}^{O\times I}$ serves as the pre-trained parameter matrix of LLMs. In LoRA, $\mathbf{A}$ and $\mathbf{B}$ are low-rank matrices with dimensions $\mathbb{R}^{r \times I}$ and $\mathbb{R}^{O \times r}$ respectively, and the rank $r$ is significantly smaller than $\min(I, O)$. The parameter $\alpha$ determines the scale of changes made to $\mathbf{W}$. Notably, when performing fine-tuning, only the matrices $\mathbf{A}$ and $\mathbf{B}$ are adjusted, leaving other components unchanged.

% where $\mathbf{x} \in \mathbb{R}^{I}$ is the representation of input information, and $\mathbf{W} \in \mathbb{R}^{{O}\times{I}}$ is the pre-trained parameter matrix of LLMs. $\mathbf{A} \in \mathbb{R}^{r \times I}$ and $\mathbf{B} \in \mathbb{R}^{O \times r}$ are the low-rank matrix in LoRA with $r \ll min(I, O)$. $\alpha$ represents the magnitude of the changes in $\mathbf{W}$. Only decomposed matrix $\mathbf{A}$ and $\mathbf{B}$ are updated in the fine-tuning process.

In instruction tuning with the task sequence in continual learning, each task is equipped with an independent LoRA expert. These experts of all tasks form a Mixture-of-Experts LoRA (i.e., \textbf{MoE-LoRA}) network. By introducing low-rank matrices, the LoRA technology greatly reduces the amount of parameter updates during model training, lowering the computational cost, while maintaining the model's learning ability.

% LoRA adopts a low-rank decomposition technique to update the parameters of the decomposed parameter matrix to learn the data distribution of the specific downstream task.
% It works on the feed-forward layer (FFN layer) in a transformer block of LLMs. For a linear layer in FFN represented as $\mathbf{h}=\mathbf{W}\mathbf{x}$, the update of the decomposed parameter matrix of LoRA is defined as:
% \begin{equation}
% \mathbf{h} = \mathbf{W}\mathbf{x} + \Delta \mathbf{W} \mathbf{x} = \mathbf{W}\mathbf{x} + \frac{\alpha}{r} \mathbf{B}\mathbf{A}\mathbf{x},
% \label{equ:lora}
% \end{equation}
% where $\mathbf{x} \in \mathbb{R}^{I}$ is the representation of input information, and $\mathbf{W} \in \mathbb{R}^{{O}\times{I}}$ is the pre-trained parameter matrix of LLMs. $\mathbf{A} \in \mathbb{R}^{r \times I}$ and $\mathbf{B} \in \mathbb{R}^{O \times r}$ are the low-rank matrix in LoRA with $r \ll min(I, O)$. $\alpha$ represents the magnitude of the changes in $\mathbf{W}$. Only decomposed matrix $\mathbf{A}$ and $\mathbf{B}$ are updated in the fine-tuning process.

\section{MoE-CL}
To address the catastrophic forgetting problem and meanwhile gain benefits from continual instruction tuning of all tasks, we propose a novel adversarial mixture of LoRA experts architecture (MoE-CL) for continual instruction tuning of LLMs. 
% Implementation details of MoE-CL are introduced in the following sections.
% As shown in Fig.~\ref{}, MoE-CL is a hybrid LoRA experts architecture, 

\subsection{Overview Architecture}
The overview architecture of MoE-CL is shown in Fig.~\ref{fig:moecl_framework}. 
It is an adversarial MoE-LoRA architecture that adaptively combines task-specific and shared experts in continual instruction tuning of LLMs. 
Each task is equipped with a dedicated LoRA expert to learn the task-specific knowledge, ensuring the parameters are independently updated so as to alleviate the catastrophic forgetting problem in continual learning. 
Meanwhile, a shared LoRA expert is designed to extract the general knowledge across all tasks to achieve high-quality cross-task knowledge transfer while minimizing interference from irrelevant information.

\subsection{Adversarial MoE-LoRA}
% As shown in Fig.~\ref{fig:moecl_framework}, MoE-CL’s adversarial LoRA architecture adaptively combines task-specific and shared experts in continual instruction tuning of LLMs. 
% Each task is equipped with a dedicated LoRA expert to learn the task-specific knowledge, ensuring the parameter independent so as to alleviate the catastrophic forgetting problem in continual learning. 
% Meanwhile, a shared LoRA expert is designed to extract the general knowledge across all tasks to achieve high-quality cross-task knowledge transfer while minimizing interference from irrelevant information.
MoE-CL employs a generative adversarial network (GAN) with a task classifier (i.e., task-aware discriminator) to explicitly constrain the parameters in task-sharing LoRA experts.

\subsubsection{The Generator in GAN}
%The generator, leveraging information encoded in the task-sharing LoRA expert, is designed to learn shared representations that deceive the task-aware discriminator.
% The input of the generator in the GAN is the shared representations in the shared LoRA expert. 
The generator in the GAN utilizes the output vector of the feed-forward layer within the transformer block as its input. The output of the generator is shared representations in the shared LoRA expert. The mathematical formulation of the generator, which transforms the feed-forward output into shared representations, is defined in Eq.~\ref{generator_function}.
% ( the generator function is formally defined in in Eq.~\ref{generator_function}).
These shared representations are designed to encapsulate common knowledge across tasks deceive the task-aware discriminator. In response, the task-aware discriminator is designed to infer the corresponding task labels from these generated shared representations.
After the training, the task-sharing LoRA expert learns high-quality cross-task sharing information, which helps the fine-tuning of subsequent tasks.
% Given the representation $\mathbf{z}_{s} \in \mathbb{R}^H$ of the shared LoRA expert,  the function of the generator is defined as:
% \begin{equation}
% xxx 
% \end{equation}

% The information in the task-sharing LoRA expert, acting as the generator within the GAN, is designed to learn shared representations without task-specific information to deceive the task-aware discriminator.
% In response, the task-aware discriminator is designed to infer the corresponding task labels from these generated shared representations.
% After the training, the task-sharing LoRA expert learns high-quality cross-task sharing information, which helps the fine-tuning of subsequent tasks.

\subsubsection{Task-aware Discriminator}
The task-aware discriminator is a task classifier to identify the label of the learning task. 
Given the input vector in $i$-th feed-forward layer in the transformer block, represented as $\mathbf{z}_i \in \mathbb{R}^H$. 
%The task-sharing LoRA expert network $F_{s}$  and the $k$-th task-specific LoRA expert network $F_{k}$ are activated for feature extraction and fusion. 
For the learning task $t$, the task-sharing representation $\mathbf{z}_{s} \in \mathbb{R}^H$ and the task-specific representation $\mathbf{z}_{t} \in \mathbb{R}^H$ are defined as follows:
\begin{equation}
    \mathbf{z}_{s} = \mathcal{F}_{LoRA} (\mathbf{z}_i, \theta_{s}),
    \label{generator_function}
\end{equation}
\begin{equation}
    \mathbf{z}_{t} =  \mathcal{F}_{LoRA}(\mathbf{z}_i, \theta_{t}),
\end{equation}
where $\mathcal{F}_{LoRA}$ is the low-rank operation function in LoRA conducted on the frozen-parameters in pre-trained LLMs.  
$\theta_{s}$ and $\theta_{t}$ are the learnable parameters in task-sharing LoRA expert and specific LoRA expert for task $t$.

% The task-aware discriminator is a task classifier to identify the label of the learning task. 
% Given the sharing representation $\mathbf{z_{s}}$ from shared LoRA expert, 
The predicted task label in the discriminator is:
\begin{equation}
    \hat{l}_t = \mathcal{F}(\mathbf{z}_{s},\phi),
\end{equation}
where $\mathcal{F}$ is the softmax activation function, and $\phi$ is the learning parameter of the task classifier.

The loss function $\mathcal{L}_{GAN}$ in GAN is computed by comparing the ground truth label $l_t$ and the predicted label from $\hat{l}_t$, defined as:
\begin{equation}
    \mathcal{L}_{GAN} = \mathcal{F}_c(\hat{l}_t, l_t),
\end{equation}
where $\mathcal{F}_c$ is the cross-entropy loss function.

\subsection{Instruction Tuning Optimization}
After instruction tuning, 
the prediction for task $t$ is computed using a weighted combination of task-sharing and task-specific representations, which are linearly interpolated with learnable weights to produce the output representation $\mathbf{z_{i+1}}$ at layer $i$
of the transformer block, defined as:
\begin{equation}
    \mathbf{z_{i+1}} = \beta_{s} \cdot \mathbf{z}_{s} + \beta_{t} \cdot \mathbf{z}_{t},
\end{equation}
% the prediction result of task $t$ is computed based on the task-sharing representation and the task-specific representation. 
% They are weighted combined to obtain the output representation of layer $i$ in the transformer block, denoted as $\mathbf{z_{i+1}}$. 

where weight coefficients $\beta{s}$ and $\beta_{t}$ of the task-sharing representation and the task-specific representation are automatically calculated by the gating network $\mathcal{G}$, formulated as:
\begin{equation}
    \beta_{s}, \beta_{t} = \mathcal{G}(\mathbf{z}_{i}).
\end{equation}
% \begin{equation}
%     \mathbf{z_{i+1}} = \beta_{s} \cdot \mathbf{z_{s}} + \beta_{t} \cdot \mathbf{z_{t}}.
% \end{equation}

The hidden representation $\mathbf{z}_{i+1}$ is then fed into a Multi-Layer Perceptron (MLP) to obtain the prediction result. The prediction loss is defined as:
\begin{equation}
    \mathcal{L}_{SFT} = \mathcal{F}_c(\hat{y}_{t}, y_t),
\end{equation}
where $y_t$ and $\hat{y}_t$ is the ground truth and predicted label respectively.
% and $\mathcal{F}_c$ represents the cross-entropy loss function.

By adding the generative adversarial loss and prediction loss in SFT, the final loss function in MoE-CL is optimized by:
\begin{equation}
    \mathcal{L} = \mathcal{L}_{SFT}-\alpha*\mathcal{L}_{GAN},
    \label{final_loss}
\end{equation}
where $\alpha$ is a positive parameter weight ranging from 0 to 1 to adjust the weight of two losses. In an ideal scenario, the discriminator can not discriminate task labels from shared representations. Hence the negative loss in GAN is used in the final optimization.

\begin{table*}[ht]
% \large
\setlength\tabcolsep{4pt}
\caption{Dataset descriptions and different orders of task sequences used in continuous learning. MTL5: a public far-domain benchmark with four text classification tasks across diverse domains; Tencent3: a real industrial dataset with content compliance review samples from three business scenarios.
% DBP, AGN, and GZPL are abbreviations for DBPedia, AGNews, and GongZhongPingLun respectively.
}
\label{tab:task_order}
\begin{tabular}{cccc|ccc}
\toprule
Benchmark        & Data &  \#Class &Task Type            & Base Model   & Order & Task Sequence       \\
\hline\hline
\multirow{3}{*}{MTL5}    & AGNews  & 4     & Topic classification   & \multirow{3}{*}{Llama 2}                                                   & 1     & DBP-\textgreater{}Amazon-\textgreater{}Yahoo-\textgreater{}AGN \\
                         & Amazon  & 5     & Sentiment analysis     &                                                                            & 2     & DBP-\textgreater{}Amazon-\textgreater{}AGN-\textgreater{}Yahoo \\
                         & DBPedia & 14    & Topic classification   &                                                                            & 3     & Yahoo-\textgreater{}Amazon-\textgreater{}AGN-\textgreater{}DBP    \\
                         & Yahoo   & 10    & Q\&A                   &                                                                            &      &     \\
\midrule
\multirow{3}{*}{Tencent3} & TASK1        & 2     & Text classification  & \multirow{3}{*}{Hunyuan}                                                 & 1     & 
                          TASK1-\textgreater{}TASK2-\textgreater{}TASK3        \\
                          & TASK2        & 2     & Text classification  &                                                                          & 2     &   TASK3-\textgreater{}TASK1-\textgreater{}TASK2        \\
                          & TASK3        & 2     & Text classification  &                                                                          & 3     & TASK2-\textgreater{}TASK3-\textgreater{}TASK1        \\
\bottomrule
\end{tabular}
\end{table*}

\section{Experiments}

\subsection{Experimental setup}

\subsubsection{Dataset}
We conduct experiments on one public continual learning benchmark \textbf{MTL5}~\cite{datasetmtl5} and a real-world large-scale industrial dataset \textbf{Tencent3}. Considering that the order of tasks in continual learning may affect the overall performance, we conduct experiments on three training orders. The details of task descriptions are shown in Table~\ref{tab:task_order}.

\begin{itemize}
% \item MTL5 is an evaluation benchmark includes five multi-classification tasks. Four of them, namely AGnews, Amazon, DBPedia, and Yahoo, are selected for continual text classification.
\item MTL5 is a far-domain benchmark that encompasses five text classification tasks across diverse domains. The significant differences between these domains render continual learning particularly challenging. We selected four tasks from this benchmark—AGnews, Amazon, DBPedia, and Yahoo—for our continual text classification experiments. Further details are provided in Table~\ref{tab:task_order}.
\item Tencent3 evaluation benchmark contains 229,442 review samples from three business scenarios, 
%i.e., ShiPinHao from WeChat Channel, XiaoShiJie from QQ, and GongZhongPingLun from WeChat Official Account, for content compliance review (i.e., binary classification task). 
% i.e., Task1 from WeChat Channel, Task2 from QQ, and Task3 from WeChat Official Account, for content compliance review (i.e., binary classification task). 
i.e., Task1 from the Review Channel, Task2 from the Social Platform, and Task3 from the Official Content in practical applications of Tencent, for content compliance review (i.e., binary classification task). 
\end{itemize}

\subsubsection{Evaluation Metrics}
We conduct comprehensive evaluations of the model's performance from three dimensions in continual learning~\cite{zheng2025towards}, namely Accuracy (Acc), Backward Transfer (BwT), and Forward Transfer (FwT). 
\begin{itemize}
\item Accuracy (Acc) is the primary indicator of model performance, reflecting the overall performance after training all tasks; higher accuracy indicates better comprehensive performance of the model.
\item Backward Transfer (BwT) evaluates the impact of subsequent tasks on previously learned tasks. A larger BwT (closer to or positive) indicates that learning new tasks has less negative impact (or even a positive impact) on old tasks, which is crucial for mitigating catastrophic forgetting.
\item Forward Transfer (FwT) measures the benefits of knowledge transfer from prior tasks to subsequent tasks. A larger FwT means more effective knowledge reuse across tasks, enhancing the model's ability to adapt to new tasks.
\end{itemize}
% As the primary indicator of model performance, Accuracy reflects the overall performance of the model after completing the training of all tasks. BWT is used to evaluate the impact of learning subsequent tasks on the performance of previously learned tasks; FwT is used to evaluate the benefits generated when knowledge is transferred from previously learned tasks to subsequent tasks.

\subsubsection{Compared Methods}.
We compare MoE-CL method with the following representative models:
\begin{itemize}
    \item \textbf{Per-task FT }: trains a separate PEFT module for each task. All tasks are independent of each other, and the training order of the tasks has no impact on the results.
    % \item \textbf{Sequential FT-F}: it directly updates all parameters of the pre-trained base LLM model, i.e., fully fine-tuning.
    \item \textbf{Sequential FT-P}: uses a shared Parameter-Efficient Fine-Tuning (PEFT) module, which is trained according to a predefined order of task sequence.
    % \item \textbf{EPI \cite{epi}}: This method assigns private delta parameters to each task and trains them in combination with the pre-trained model. 
    % % During testing, it uses a non-parametric task identification method to select appropriate parameters for prediction and promotes knowledge sharing between tasks through a specific approach.
    \item \textbf{O-LoRA~\cite{olora}}: learns tasks in orthogonal low-rank vector subspaces to minimize interference of each other in continual learning.
    \item \textbf{MoCL~\cite{wang-etal-2024-rehearsal}}: is the current state-of-the-art method. It equips each task with a dedicated PEFT module and calculates the similarity between the input and the task vector as weights to fuse the outputs of the training task.
    \item \textbf{MoE-CL}: our adversarial MoE architecture, in which a dedicated LoRA expert is used for each task to preserve task-specific knowledge and a shared LoRA expert to facilitate cross-task knowledge transfer.
\end{itemize}

\subsubsection{Implementation Details.}
% All methods are implemented in PyTorch with an 8-GPU H20 environment.
% For each baseline method, a grid search is applied to find the optimal settings. 
% These include learning rate from ranging from 0.0001 to 0.001, taking values at intervals of 0.0001. 
% And rank of the LoRA matrix from $\{2, 4, 8, 16, 32\}$.
% We report the results of each method with its optimal hyperparameter settings on the validation data. 
% In our model, the dimension $H$ of the task-shared and task-specific vectors is the size of the FFN output layer in the base model.
% The learning rate is 0.0002, the rank of LoRA matrix is set to 8 in MTL5 and Tencent3 benchmarks. 
% The balance weight $\alpha$ in Eq.~\ref{final_loss} is set to 0.1.
% The implementation code will be available after the review process.  

All methods are implemented in PyTorch within an 8-GPU H20 environment. For each baseline method, grid search is conducted to determine optimal hyperparameters, including learning rates ranging from 0.0001 to 0.001 (stepping by 0.0001) and LoRA matrix ranks in $\{2, 4, 8, 16, 32\}$. Results reported for each method use validation-data-optimized hyperparameters.  
In our model, the dimension $H$ of task-shared and task-specific vectors matches the FFN output layer size of the base model, which is 4096. For MTL5 and Tencent3 benchmarks, the learning rate is set to 0.0002, the LoRA matrix rank to 8, and the balance weight $\alpha$ in Eq.~\ref{final_loss} to 0.1.

\begin{table*}[]
% \large
\centering
\caption{Evalutions on Tencent3 evaluation benchmark conducted on \textbf{Tencent Hunyuan} foundation model. The \emph{underline} represents the SOTA compared method according to the primary evaluation metric Accuracy. \emph{Bold} font indicates that the model has the best comprehensive performance in terms of both Accuracy and Stability.}
\label{tab:tencent3_results_hunyuan}
\begin{tabular}{c|c|lllll}
\toprule
\multirow{2}{*}{Metric} & \multirow{2}{*}{Order} & \multicolumn{5}{c}{Method}                                                                \\
                        &                        & Per-task   FT   & Sequential FT-P & O-LoRA           & MoCL            & \textbf{MoE-CL}           \\
\hline \hline
\multirow{2}{*}{Accuracy ($\uparrow$) }   & Avg                   & 0.5334         &  \emph{0.6071$\pm${0.0220}}          & 0.5950$\pm$0.0122           & 0.5918$\pm$0.0293          &
                        \textbf{0.6342$\pm$0.0074}           \\
                     & 1                      & 0.5334          & 0.6365          & 0.5901           & 0.5764          & 0.6446           \\
                 \textbf{(Primary Indicator)}          & 2                      & 0.5334          & 0.6012          & 0.6118           & 0.6328          & 0.6280                    \\
                        & 3                      & 0.5334          & 0.5836          & 0.5832           & 0.5663          & 0.6299           \\
\midrule
\multirow{4}{*}{BwT ($\downarrow$) }    & Avg                   & -0.1593         & \emph{-0.0300$\pm$0.0324}         & -0.0223$\pm$0.0024          & -0.0485$\pm$0.0249         & 
                    \textbf{-0.0349$\pm$0.0168}           \\
                        & 1                      & -0.1593         & -0.0632         & -0.0192          & -0.0747         & -0.0578          \\
                        & 2                      & -0.1593         & -0.0406         & -0.0251          & -0.0150         & -0.0289          \\
                        & 3                      & -0.1593         & 0.0139          & -0.0225          & -0.0559         & -0.0179          \\
\midrule
\multirow{4}{*}{FwT  ($\uparrow$)}    & Avg                    & 0.0562          & \emph{0.0578$\pm$0.0287}          & 0.0106$\pm$0.0078           & -0.0139$\pm$0.0052         & 
                        \textbf{0.0573$\pm$0.0159}           \\
                        & 1                      & 0.0562          & 0.0916          & -0.0003          & -0.0098         & 0.0797           \\
                        & 2                      & 0.0562          & 0.0603          & 0.0173           & -0.0212         & 0.0485           \\
                        & 3                      & 0.0562          & 0.0214          & 0.0147           & -0.0106         & 0.0438           \\ \hline \hline
                       % \multicolumn{2}{c|}{Latency (ms/sample)} & 103.71 & 216.05 & 106.02 & 107.11 & 144.46 \\
                       \multicolumn{2}{c|}{\textbf{Latency} (ms/sample)} & 4.5ms & 9.4ms & 4.6ms & 4.7ms & 6.3ms \\
\bottomrule
\end{tabular}
\end{table*}

\begin{table}[]
% \large
\centering
\caption{The Accuracy on the MTL5 benchmark. %The bold text represents the best performance.
}
\label{tab:mtl5_acc}
    \begin{tabular}{ccccc}
        \toprule
        \multirow{2}{*}{MTL5 (Llama 2)} &  \multirow{2}{*}{Avg} & \multicolumn{3}{c}{ Orders}                        \\ 
                                &        & 1             & 2             & 3             \\
        \hline \hline
        Sequential FT-P         & 26.7$\pm0.91$          & 28.8          & 27.4          & 26.6          \\
        Per-task FT             & 76.6$\pm0.00$          & 76.6          & 76.6          & 76.6          \\
        O-LoRA                  & 76.1$\pm0.52$          & 76.8          & 75.7          & 75.7          \\
        MoCL                    & 78.2$\pm0.33$          & 78.4          & 77.7          & 78.4          \\
        \textbf{MoE-CL(Ours)}            & \textbf{80.5$\pm$1.50} & \textbf{81.1} & \textbf{81.9} & \textbf{78.4} \\
        \bottomrule
    \end{tabular}
\end{table}

\subsection{Main Results}
The experimental results on MTL5 and Tencent3 evaluation benchmarks are shown in Table \ref{tab:mtl5_acc} and Table~\ref{tab:tencent3_results_hunyuan}. We have the following observations:

% MTL5-Acc

(1) 
Our proposed method MoE-CL demonstrates remarkable improvements in average accuracy (\textbf{Avg.ACC}) compared to all baseline approaches with minimal variance, highlighting its superior generalization performance and robust stability across diverse task complexities.
% by independently learning knowledge from the task-specific expert and filtering the task-irrelevant information by sharing experts based on the task discriminator. 

(2)  % 讲 两个benchmark 中最差的方法，分析原因，如 mt 5 任务差距大？等
Sequential FT-P exhibits inconsistent performance across two benchmarks: it achieves the worst \textbf{Avg.ACC} in the MTL5 benchmark due to its parameter-sharing strategy exacerbating catastrophic forgetting on highly heterogeneous tasks (e.g., topic classification, Q\&A, sentiment analysis) with significant semantic gaps. Conversely, on the Tencent3 benchmark, which comprises homogeneous content compliance review classification tasks, it achieves the second-best performance after our MoE-CL.
% Sequential FT-P demonstrates the worst average accuracy (\textbf{Avg.ACC}) in the MTL5 benchmark due to its parameter-sharing strategy exacerbating catastrophic forgetting on higher heterogeneous tasks (e.g., topic classification, Q\&A, Sentiment analysis) with significant semantic gaps. In contrast, on the Tencent3 benchmark (which consists of homogeneous content compliance review classification tasks), Sequential FT-P achieves the second-best performance after our MoE-CL. 
% The lack of task-type diversity allows Sequential FT-P to effectively transfer shared knowledge across tasks through parameter sharing. 
% Per-task FT performs poorest on Tencent3 as its independent optimization for each task ignores shared representations, failing to exploit commonalities among similar tasks.

% In the MTL5 benchmark, Sequential FT-P has the worst average performance. This is because multiple tasks jointly optimize the same PEFT module, causing task interference that impacts the final result.
% In the Tencent3 benchmark, Per-task FT performs the worst. It optimizes a PEFT module for each task independently, missing out on positive knowledge transfer between tasks. Additionally, on Tencent3, Sequential FT-P ranks second only to our MoE-CL, a significant contrast to its MTL5 performance. This is due to the smaller task gaps on Tencent3, providing much shared knowledge for Sequential FT-P to transfer.

(3) % tencent 中BWT 和 FWT 在各个方法中的表现, 也可以分成两个小点分别说两个指标
%　For \textbf{BWT} and \textbf{FwT} evaluation metrics, 
%which are used to evaluate the impact of learning subsequent tasks on the performance of previously and later learned tasks, respectively. 
% From the results on Tencent3 evaluation benchmark,
% we can see that 
Our model achieves fewer negative effects from subsequent tasks than MoCL (BwT) and 
exhibits stronger \textbf{Stability} than Sequential FT-P on BwT and FwT, showing the effectiveness of integrating hybrid LoRA experts in a GAN-based architecture.
% which reduces the forgetting of task-specific knowledge and enhances model performance by sharing knowledge from previous tasks.
O-LoRA performs best in the BwT metric because it reserves an orthogonal low-rank parameter space for each task that avoids interference from subsequent tasks. 
% In comparison, our proposed method shows fewer negative effects from subsequent tasks than MoCL and exhibits stronger stability than Sequential FT-P on FwT, showing the effectiveness of integrating task-specific LoRA experts, which reduces the forgetting of previous knowledge and enhances overall model performance.

% (4) % FwT 分析 
% As for FwT, which is used to evaluate the benefits generated when knowledge is transferred from previously learned tasks to subsequent tasks.  Sequential-FT-P performs best in the FwT metric. As it constantly optimizes the same PEFT module, it can transfer knowledge to the greatest extent. Moreover, the tasks in Tencent3 have small differences, providing a lot of positive knowledge for transfer.
% Our proposed MoE-CL ranks second, indicating that well-trained shared experts can also efficiently transfer cross-task shared knowledge.

(4) % task sequence 会影响结果，但是我们的方法更加鲁棒， 通过分离的 share and specific experts
On the MTL5 and Tencent3 benchmarks, task sequence order significantly impacts model performance.
Our method, MoE-CL, demonstrates superior \textbf{Stability} across different task sequence orders through its architecture of explicitly separating shared experts (handling common knowledge across tasks) and specific experts for each task.
% This design, combined with the generative adversarial network, effectively transfer useful knowledge and minimize the forgetting problem in continual learning process. 
In contrast, methods like MoCL (relying on task vector similarity) and Sequential FT-P (suffering from catastrophic forgetting) showed greater sequence-order sensitivity.

(5) A significant improvement and stability have been achieved by MoE-CL in continual learning, but its implementation involves some architectural complexity. We test the inference \textbf{Latency} of our model in an 8-GPU H20 environment. Compared to other models, our model has a relatively higher latency, i.e., 6.3ms for a sample (with avg.length of 300 tokens), but it remains within the range imperceptible to humans and is acceptable in real industrial scenarios.

\begin{figure}
  \centering
  \includegraphics[width=0.5\textwidth]{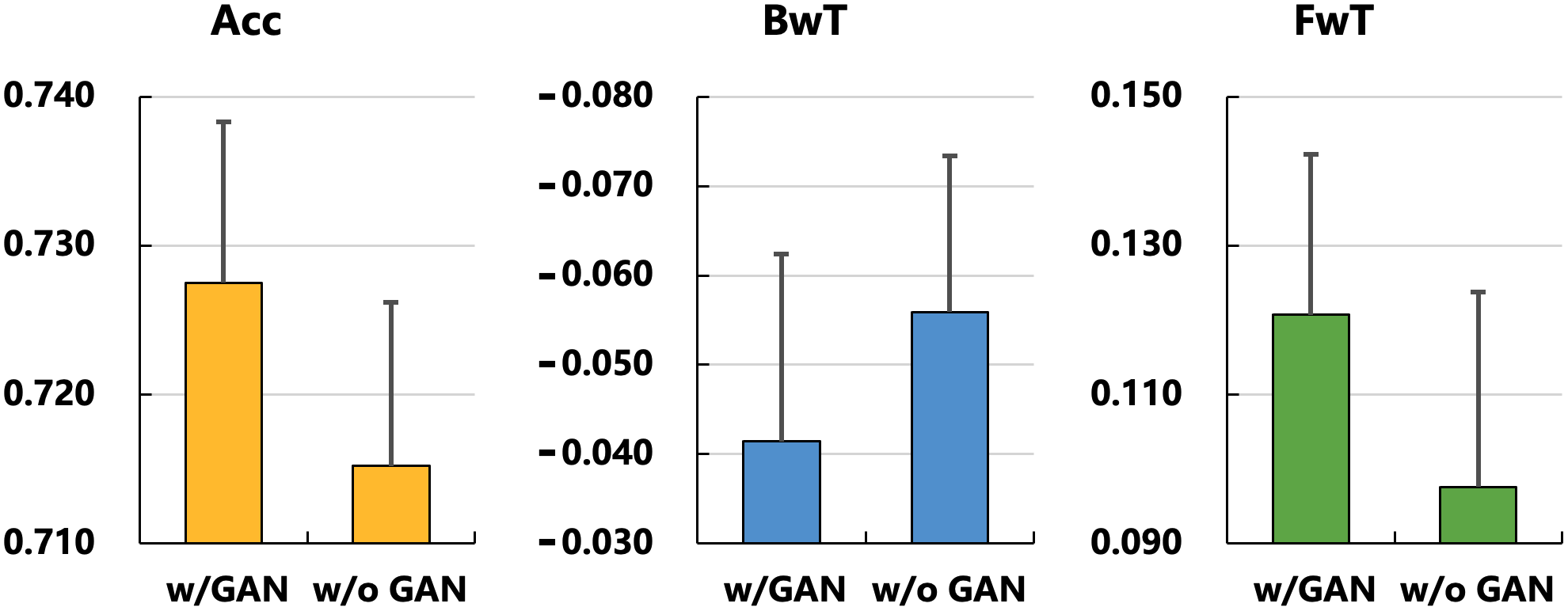}
  \caption{The impact of GAN-based architecture in MoE-CL: Performance comparisons (on Accuracy, Backward Transfer, and Forward Transfer) between MoE-CL with GAN (suppressing task-irrelevant noise in shared expert) and without GAN, demonstrating GAN's role in alleviating catastrophic forgetting and enhancing cross-task generalization in continual instruction tuning on MTL5 benchmark.}
  \label{fig:influ_gan}
\end{figure}

\begin{table}
\small
\centering
\caption{Offline A/B test measured by \textbf{Stripping Rate}. MoE-CL reduces 15.3\% and 3.2\% manual review manpower costs in real business scenarios on a Video Platform and Social Platform in Tencent Security.
}
\label{tab:offline_ab_test}
\begin{tabular}{c|ccc|ccc}
% \toprule
% \multicolumn{7}{c}{Stripping Rate}                                                                                                   \\
\toprule
\multirow{2}{*}{Method} & \multicolumn{3}{c|}{Video Platform}                       & \multicolumn{3}{c}{Social Platform}                       \\
                        & Class 0         & Class 1         & \textbf{SUM}             & Class 0          & Class 1         & \textbf{SUM}             \\
\hline \hline
Online                  & 5.3\%           & 8.2\%           & 13.5\%          & 22.2\%           & 12.0\%          & 34.2\%          \\
MoE-CL                  & 15.5\%          & 13.0\%          & 28.8\%          & 20.6\%           & 16.8\%          & 37.4\%          \\
\midrule
\textbf{Gain}                    & \textbf{10.2\%$\uparrow$}  & \textbf{4.8\% $\uparrow$}  & \textbf{15.3\% $\uparrow$} & -1.6\% $\downarrow$           & \textbf{4.8\% $\uparrow$}  & \textbf{3.2\%$\uparrow$}  \\
\bottomrule
\end{tabular}
\end{table}

% \subsection{Experimental Analysis}
\subsection{Ablation Study}

To verify the effectiveness of our adversarial MoE-LoRA architecture in addressing the catastrophic forgetting problem, we remove the GAN component in the MoE-CL (i.e., w/o GAN). The comparisons with the degraded version are shown in Fig.~\ref{fig:influ_gan}, which contrasts performance across three key metrics for continual learning: Accuracy (Acc, overall task performance), Backward Transfer (BwT, evaluating the impact of subsequent tasks on prior ones), and Forward Transfer (FwT, measuring knowledge transfer benefits from prior to new tasks). For the three metrics, higher values are indicative of superior performance. As for BWT, it is generally negative, but less negative values (closer to 0) indicate less negative impacts: the model performs better in addressing the catastrophic forgetting problem.

We observe that our GAN-based MoE expert architecture achieves a notable improvement in the average performance across all tasks in CL: it outperforms the w/o GAN variant in Accuracy (with higher overall scores), exhibits less negative BwT (indicating reduced interference of new task training on old tasks), and yields higher FwT (showing more effective cross-task knowledge reuse). Our GAN-based MoE architecture effectively localizes task-specific information within task-specific LoRA experts by suppressing task-irrelevant noise in the shared expert through adversarial training, ensuring that only task-aligned knowledge is transferred across tasks. This strategic distribution prevents the occurrence of catastrophic forgetting, a phenomenon clearly reflected in the BwT metric—where the GAN-integrated model shows significantly smaller performance degradation on previously learned tasks after training new ones compared to the version without GAN.

\subsection{Offline A/B Testing}

In \emph{Content Compliance Review} (\textbf{CCR}) in Tencent Security, content samples with machine-predicted confidence scores exceeding a predefined threshold are classified into two types.
\begin{itemize}
\item \textbf{Class 0: white}. Denoting the compliant content that meets regulations (automatically approved via high-confidence model predictions).
\item \textbf{Class 1: black}. Denoting non-compliant content that violates rules (automatically flagged or blocked via high-confidence predictions).
\end{itemize}
Either "Class 0: white" (compliant) or "Class 1: black" (non-compliant), bypassing manual human review. 
The effectiveness of applied algorithms is measured by the \textbf{Stripping Rate}, which quantifies the proportion of high-confidence samples automatically classified without manual review.
The higher stripping rate reduces operational costs by minimizing the volume of content requiring human intervention.
% The effectiveness of this process is measured by the \textbf{Stripping Rate}, which 
% quantifies the proportion of high-confidence samples automatically classified without manual review.
We conduct offline A/B testing to compare the stripping rate of our proposed MoE-CL method against the applied production algorithm. 
As presented in Table~\ref{tab:offline_ab_test}, 
MoE-CL achieved significant stripping rate improvements in the CCR task across both the Video Platform and Social Platform practical applications in Tencent. 
%  ShiPinHao and XiaoShiJie scenarios. 
% In particularly, it attained a stripping rate of 28.8\% in ShiPinHao (15.3\% higher than the production baseline), translating to an 15.3\% reduction in manual review manpower costs, delivering tangible business value through operational efficiency gains.
In the Video Platform scenario, it achieves a 15.3\% improvement in stripping rate over the production baseline. This uplift directly reduced manual-review manpower costs by 15.3\%, delivering tangible business value through operational efficiency gains and highlighting the solution’s impact in real-world industrial applications.

% MoE-CL achieved a stripping rate in CCR task of 28.8\% in ShiPinHao and 37.4\% in XiaoShiJie, representing improvements of 15.3\% and 3.2\% respectively over the production baseline. 
% This performance leap directly reduced manual review manpower costs by 15.3\% and 3.2\% respectively, delivering tangible business value through operational efficiency gains.

% In Tencent's actual business scenarios, when the probability predicted by the machine review for a sample exceeds a certain threshold, human review can be skipped. This approach aims to reduce the volume of data that requires human review and further cut down on the manpower needed for the review process. The reduction in the amount of data for human review is quantified by the stripping rate, which is calculated independently for white (compliant) and black (non-compliant) samples.
% Specifically, for white samples, the stripping rate is determined as follows: First, a threshold is set so that the precision rate of white samples is greater than 0.9. Then, the stripping rate of white samples is calculated by dividing the number of samples predicted as white by the total number of samples under this threshold. The calculation method for black samples is similar. Samples that fall outside the confident prediction ranges for both types still need further human review.
% As a result, a higher stripping rate indicates better model performance and greater savings in the manpower costs associated with the review process.

\section{Conclusion}
% We introduce MoE-CL, a novel Mixture of LoRA Experts architecture designed to address catastrophic forgetting and enable effective knowledge transfer in large language models during continual instruction tuning. By combining dedicated LoRA experts for task-specific knowledge retention with shared LoRA experts augmented by a GAN-based task-aware discriminator, MoE-CL achieves a balance between preserving prior task performance and incorporating new task knowledge. Experiments on Tencent’s real-world content compliance review systems demonstrate the effectiveness of MoE-CL, validating its practical viability in large-scale industrial AI deployment.
% While MoE-CL significantly advances continual learning for LLMs, its current implementation faces architectural complexity and hyperparameter tuning challenges. Despite these constraints, it maintains comparable inference speed and easier convergence in real-world Tencent scenarios, which we plan to optimize in our future work.
We introduce MoE-CL, a novel Mixture of LoRA Experts architecture that effectively addresses catastrophic forgetting and enables robust knowledge transfer in large language models during continual instruction tuning—key to enabling the self-evolution of LLMs. By integrating dedicated LoRA experts for task-specific knowledge retention with shared LoRA experts enhanced by a GAN-based task-aware discriminator, MoE-CL not only achieves a remarkable balance between preserving prior task performance and incorporating new task knowledge but also fosters the self-evolution of LLMs by enabling autonomous adaptation to sequential tasks without heavy external intervention. Our experiments on Tencent’s real-world content compliance review systems not only demonstrate the effectiveness of MoE-CL but also validate its practical viability in large-scale industrial AI deployment with sustained self-evolution capabilities.

While MoE-CL represents a significant advancement in enabling LLM self-evolution through continual learning, we acknowledge that its current implementation involves some architectural complexity. However, these challenges do not detract from its core strengths. MoE-CL maintains a comparable inference speed and achieves easier convergence in real-world scenarios at Tencent, highlighting its potential for optimization and scalability in supporting sustained self-evolution. Future work will focus on refining these aspects to further enhance its self-evolution capabilities and broaden its applicability.

\begin{acks}
This work is sponsored by the CCF-Tencent RAGR20240108, the National Natural Science Foundation of China under Grant No. 62236003, and the Tencent Basic Platform Technology Rhino-Bird Focused Research Program.
\end{acks}

\bibliographystyle{ACM-Reference-Format}
\bibliography{sample-base}

%%
%% If your work has an appendix, this is the place to put it.
\appendix

\end{document}